\documentclass[11pt,a4paper]{article}
\usepackage[hyperref]{emnlp2018}
\usepackage{times}
\usepackage{latexsym}
\usepackage[utf8]{inputenc}

\usepackage{url}
\usepackage{graphicx}
\usepackage{multicol}
\usepackage{multirow}
\usepackage{float}

\aclfinalcopy 

\title{Monolingual and Cross-lingual Zero-shot Style Transfer}

\author{Elizaveta Korotkova \hspace{5mm} Maksym Del \hspace{5mm} Mark Fishel \hspace{5mm} \\
  Institute of Computer Science\\
  Universitu of Tartu, Estonia \\
  {\tt \{elizaveta.korotkova,maksym.del,fishel\}@ut.ee}}

\date{}

\begin{document}
\maketitle

\begin{abstract}

We introduce the task of zero-shot style transfer between different languages. Our training data includes multilingual parallel corpora, but does not contain any parallel sentences between styles, similarly to the recent previous work. We propose a unified multilingual multi-style machine translation system design, that allows to perform zero-shot style conversions during inference; moreover, it does so both monolingually and cross-lingually. Our model allows to increase the presence of dissimilar styles in corpus by up to 3 times, easily learns to operate with various contractions, and provides reasonable lexicon swaps as we see from manual evaluation.

\end{abstract}

\section{Introduction}
It is crucial for intelligent natural language generation systems to produce text which is appropriate for the task at hand in terms of semantic content as well as style. 

Due to the shortage of parallel corpora which could be used for learning to transfer style in a supervised fashion, much of recent work has been focusing on style transfer using non-parallel data \cite{mt-backtrans, gvae-2, gvae-3}. The proposed approaches mostly involve learning style representations disentangled from semantic content. Other work has been based on paraphrase data \cite{rnns-zeroshot} or explicit attribute substitution \cite{manual}.

We do not rely on explicit separation between content and style representations nor on explicit attributes. Instead, we approach the problem from a different angle.

In this paper we propose a multilingual multistyle machine translation system that allows to modify stylistic traits of a sentence while also translating it into a different language. It uses ideas from multilingual NMT \cite{multiling-goolge, multiling-multiway}, while extending them to styles, and relies on source word factors as a key design choice \cite{factors}.
We use no parallel data between styles, but still can perform cross-style conversions during inference; moreover, we do so cross-lingually.

In our experiments, we evaluate the proposed model incorporating three languages and three styles. The system shows strong results as a pure machine translation model as well, which provides for good meaning preservation and fluency in output texts. The system can be easily expanded to incorporate more languages and styles (though scaling up to more languages and styles is to be tested).

The paper contains the following contributions:
\begin{itemize}
\item  Extends the task of style transfer to multiple languages resulting in cross-lingual style transfer
\item  Tackles the task of cross-lingual style transfer without parallel data between styles
\item  Proposes a unified multilingual multi-style system design that enables cross-lingual zero-shot style transfer   
\end{itemize}

This paper is organized as follows. In Section \ref{sec:problem}, we formally state the problem we attempt to solve, that is, cross-lingual zero-shot style transfer. We then describe our general approach in Section \ref{sec:approach} and specific details of the experiments, their results, and evaluation in Section \ref{sec:experiments}. Some discussion points are put forward in Section \ref{sec:discussion}. Related work in the field is summarized in Section \ref{sec:related-work}. Finally, in Sections \ref{sec:conclusion} and \ref{sec:future} we present our conclusions and plans for future work.

\section{Problem Statement}
\label{sec:problem}

Let $L=\{l_1, l_2, ..., l_m\}$ be a set of languages 
and $S=\{s_1, s_2, ..., s_n\}$ a set of text styles\footnote{We treat the term ``text style'' loosely as covering concepts like text domain, genre, formality and other text characteristics.}.

Let $e(l_x, s_y)$ and $e(l_q, s_p)$ be sentences representing the same semantic content, but in different languages ($l_{x/q}$) and styles ($s_{y/p}$): $x,q \in \{1..m\}$, and $y,p \in \{1..n\}$, while $x{\neq}q$ and $y{\neq}p$.

\begin{figure}[]
    \includegraphics[width=3.0in]{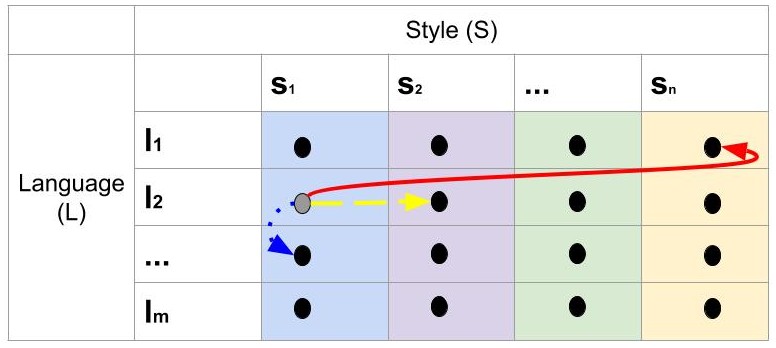}
  \caption{\small Machine Translation and Style Transfer tasks. Blue (dotted) line describes the usual Machine Translation task which is generally done inside a single domain. Yellow (dashed) line corresponds to Monolingual Style Transfer task. Red (solid) line represents Cross-lingual Style Transfer Task. If parallel data is available only between multiple languages inside one style (highlighted columns), and not between styles, then the red (solid) and yellow (dashed) lines correspond to Zero-shot Transfer Tasks.
      }
  \label{fig-problem}
\end{figure}

Our task then is to 
learn the mapping between sentences like 
$e(l_x, s_y)$ and $e(l_q, s_p)$, which we call Cross-lingual Style Transfer. Refer to Figure \ref{fig-problem} for more details and comparison to tasks of Monolingual Style Transfer and Machine Translation.


A key point is that we do not use any parallel data between styles neither monolingually inside a single language nor cross-lingually. Since the only data available is the data between different languages inside individual styles, the task becomes harder and result predictions can be called "Zero-shot".

\section{Approach}
\label{sec:approach}

For our attempt at cross-lingual style transfer, we train a factored multilingual neural machine translation system, passing the desired target language and style as factors (word features), token-parallel to source texts. 

The source side of one training example $e$, where we translate source words $e_1, e_2, ..., e_n$ into language $l_{tgt}$ and style $s_{tgt}$, looks like following: 
$$
| \>e_1 \>|\> l_{tgt} \>|\> s_{tgt} \>| \quad 
| \>e_2 \>|\> l_{tgt} \>|\> s_{tgt} \>| \quad
...  \quad
|\> e_N \>|\> l_{tgt} \>|\> s_{tgt} \>|,
$$
where vertical lines highlight input words factors.  The target sentence remains unchanged.

\begin{figure}[t]
    \includegraphics[width=3.0in]{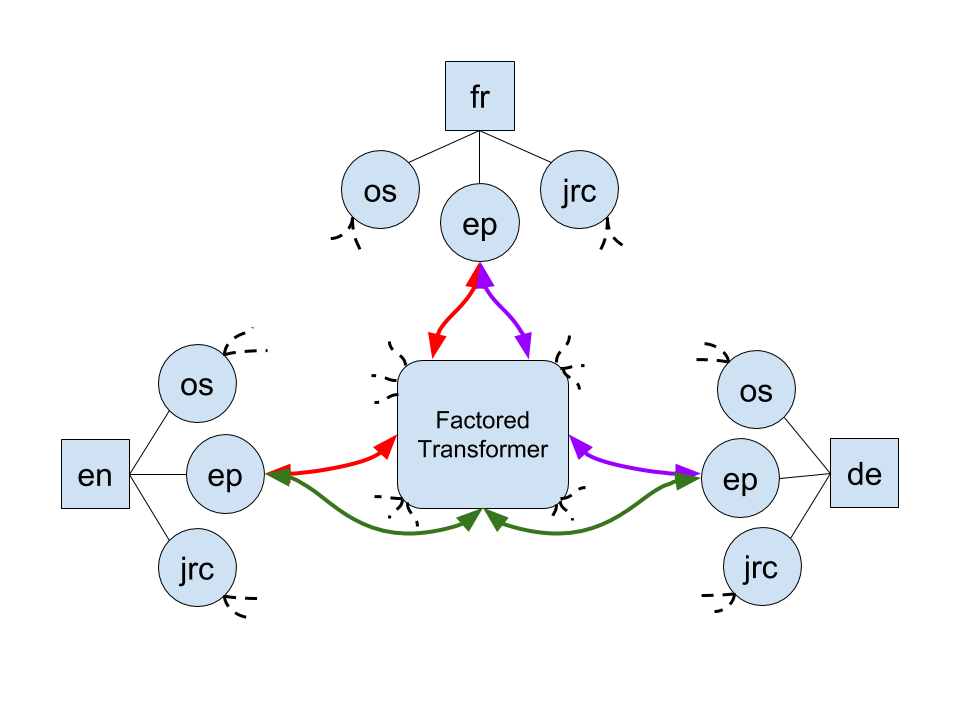}
  \caption{\small Proposed Multilingual Multistyle Machine Translation System. In this experiment, we consider 3 languages: English (en), Franch (fr), German (de), and 3 styles: OpenSubtitles (OS), Europarl (EP), and JRC-Acquis (JRC). The image shows the data flow between language-style pairs. Arrows of the same color correspond to the parallel data. During training, source sentences in a particular language and belonging to a particular style are encoded with Factored \cite{factors} Transformer \cite{transformer} model and decoded into a different language, but the same style. At inference time, we are able to convert source sentences into different languages, and, importantly, into different styles (despite not having cross-style parallel data for training). Information about languages and styles is communicated to the model in form of the word factors.}
  \label{fig-dataflow}
\end{figure}

See Figure \ref{fig-dataflow} for a concrete example of the training architecture and data flow.
In this way, during training the model encounters examples of translation in different language directions, but texts belonging to a certain style are only translated into the same style, since we do not use any style-parallel data. Our assumption is that given enough data for different languages and styles the model should learn to match the translations not just to the output language but also to the desired style, even without having seen style-wise parallel data during training.

To perform cross-lingual style transfer, we translate texts using the trained model, passing the desired target language and style, which now may be different from the style of the source text, as factors.

\section{Experiments}
\label{sec:experiments}

\subsection{Datasets}

The translation model was trained using three parallel corpora: OpenSubtitles2018 \footnote{\url{http://www.opensubtitles.org/}} , a new release of the corpus presented by
\citet{opensubs}, Europarl \citep{europarl} and JRC-Acquis \citep{jrc}. OpenSubtitles2018 is a corpus made up of movie and TV subtitles, Europarl consists of texts of European Parliament proceedings, and JRC-Acquis comprises legislative texts of the European Union.

We assume that there should be sufficient stylistic difference between these three corpora, especially between the more informal OpenSubtitles2018 and the other two, while acknowledging the fact that most text corpora and OpenSubtitles in particular constitute a heterogeneous mix of genres and text characteristics; however, many stylistic traits are also similar across the whole corpus, which means that these common traits can be learned as a single style.

We used three language pairs: English-German, English-French and German-French, each taken in both directions. Refer to the Figure \ref{fig-dataflow} for the system architecture and data flow.

Prior to model training, we discarded all sentence pairs where at least one sentence  was an empty string, consisted of more than 100 tokens, or did not contain any alphabetic characters, and pairs where token-wise length ratio exceeded 9. Table \ref{tab-trainsizes} shows the sizes of the training sets.

\begin{table}[]
\centering
\begin{tabular}{|c|c|c|c|}
\hline
                                     & OS & EP    & JRC   \\ \hline
de $\leftrightarrow$ en & 3M & 1.95M & 0.71M \\ \hline
en $\leftrightarrow$ fr & 3M & 2.04M & 0.71M \\ \hline
de $\leftrightarrow$ fr & 3M & 1.93M & 1.47M \\ \hline
\end{tabular}
\caption{\small Training set sizes (number of sentence pairs).
      }
      \label{tab-trainsizes}
\end{table}

\subsection{Experiment Details}

All corpora were preprocessed in a standard way: they were tokenized using Moses tokenizer ~\citep{moses}, true-cased and, finally, segmented using the SentencePiece\footnote{\url{https://github.com/google/sentencepiece}} \citep{googleSentencePieces}
implementation of sub-word units with a joint vocabulary of size 32000.

We trained an NMT model using the Sockeye framework ~\citep{sockeye}. We used transformer encoder and decoder with 6 layers, transformer model size of 512, 8 attention heads and ReLU activation for both encoder and decoder. Adam optimizer was used. Source and target token embeddings were both of size 512, and factors determining target language and style had embeddings of size 4. Batch size was set to 2048 words, initial learning rate to 0.0002, reducing by a factor of 0.7 every time when validation perplexity has not improved for 8 checkpoints, which happened every 4000 updates.

The model finished training during the 11th epoch, when validation perplexity has not improved for 32 consecutive checkpoints.

The parameters of a single best checkpoint were used for all translations, with beam size set to 5.

For evaluation we use the BLEU \citep{bleu} and METEOR \citep{meteor} metrics in cases where parallel test data is available. For evaluating style transfer without parallel data we perform a qualitative comparison using a style classifier, described in the following section.

\subsection{Evaluation Results}

When prompted to translate sentences into different styles, the model shows the ability to capture some key characteristics of those styles. One such characteristic in the English language is the presence of contracted forms, such as \textit{I'll}, \textit{she's} and so on. Contracted forms are typically abundant in informal texts, such as those making up the OpenSubtitles corpus, and are typically absent in formal and official texts, such as parliament proceedings in Europarl and legal documents in JRC-Acquis. Tables \ref{tab-deencontr} and \ref{tab-frencontr} show the number of contractions in test sets translated, respectively, from German and French into English in all possible style directions.

\begin{table}[]
\centering
\begin{tabular}{|c|c|c|c|}
\hline
\multirow{2}{*}{$S_{src}$} & \multicolumn{3}{c|}{$S_{tgt}$}\\
\cline{2-4} 
                      & OS                 & EP             & JRC            \\ \hline
OS                    & \textbf{462 (350)} & 14             & 25             \\ \hline
EP                    & 320                & \textbf{0 (0)} & 0              \\ \hline
JRC                   & 31                 & 0              & \textbf{0 (0)} \\ \hline
\end{tabular}
\caption{\small Number of contractions in 1000-sentence test sets when translated from German into English in all 9 style directions. The numbers in parentheses indicate the number of contracted forms in the human-translated test sets.
      }
      \label{tab-deencontr}
\end{table}

\begin{table}[]
\centering
\begin{tabular}{|c|c|c|c|}
\hline
\multirow{2}{*}{$S_{src}$} & \multicolumn{3}{c|}{$S_{tgt}$}\\
\cline{2-4} 
                      & OS                 & EP             & JRC            \\ \hline
OS                    & \textbf{469 (352)} & 10             & 11             \\ \hline
EP                    & 327                & \textbf{0 (1)} & 0              \\ \hline
JRC                   & 35                 & 0              & \textbf{0 (0)} \\ \hline
\end{tabular}
\caption{\small Number of contractions in 1000-sentence test sets when translated from French into English in all 9 style directions. The numbers in parentheses indicate the number of contracted forms in the human-translated test sets.
      }
      \label{tab-frencontr}
\end{table}

One other aspect that the model captures is the different lexical and grammatical choices appropriate for different styles. Some examples of different wording that the system uses when translating the same source sentences into different domains can be found in Table \ref{tab-deenexamples} for German-English and in Table \ref{tab-frenexamples} for French-English. For more and longer examples refer to Supplemental Material \ref{sec:supplemental}.

\begin{table}[]
\small
\centering
\begin{tabular}{|p{1.5cm}|p{1.5cm}|p{1.5cm}|p{1.5cm}|}
\hline
\textbf{Source (de)}      & \textbf{OS}             & \textbf{EP}                     & \textbf{JRC}               \\ \hline
m{\"o}chte ich       & i want         & i would like           & i would like      \\ \hline
aber             & but            & however                & however           \\ \hline
ich wechsle      & i'm turning    & i am turning           & i shall refer     \\ \hline
tats{\"a}chlich      & actually       & indeed                 & indeed            \\ \hline
wir gehen        & we're leaving  & we are leaving         & we shall leave    \\ \hline
noch viel mehr   & more that than & much more              & even more so      \\ \hline
lindern          & ease           & alleviate              & alleviate         \\ \hline
ja               & yeah           & yes                    & yes               \\ \hline
chance           & chance         & opportunity            & opportunity       \\ \hline
bleibt           & stay           & remain                 & remain            \\ \hline
wie schon gesagt & like i said    & as i have already said & as already stated \\ \hline
au{\ss}erdem         & besides        & moreover               & moreover          \\ \hline
ach ja?          & oh, yeah?      & is that so?            & is that so?       \\ \hline
wir gehen          & we're leaving      & we are leaving            & we shall leave       \\ \hline
wie geht's dann weiter?          & so, what's next?      & what happens then?            & how are we to proceed then?       \\ \hline
\end{tabular}
\caption{\small Examples of different wording in German-English style transfer. The German phrase \textit{m{\"o}chte ich} will be translated into OpenSubtitles as \textit{I want}, and into Europarl and JRC-Acquis as \textit{I would like}. \textit{lindern} and \textit{bleiben} become \textit{ease} and \textit{stay}, respectively, in OpenSubtitles, but \textit{alleviate} and \textit{remain} in Europarl and JRC-Acquis. In some other cases there is also difference between the formal styles: \textit{wie geht's dann weiter?} becomes \textit{so, what's next?} in OpenSubtitles, \textit{what happens then?} in Europarl and \textit{how are we to proceed then?} in JRC-Acquis.
      }
      \label{tab-deenexamples}
\end{table}

\begin{table}[]
\small
\begin{tabular}{|p{1.5cm}|p{1.5cm}|p{1.5cm}|p{1.5cm}|}
\hline
\textbf{Source (fr)}                  & \textbf{OS}                      & \textbf{EP}                       & \textbf{JRC}                     \\ \hline
salut                   & hey                     & hi                       & hi                      \\ \hline
on se lance?            & let 's go.             & are we getting started? & are we going?          \\ \hline
un d{\'e}lai de deux heures & two hours               & two hours                & a two-hour period       \\ \hline
c'est {\c c}a              & that 's right         & that is it             & that is the case      \\ \hline
vrai                    & real                    & genuine                  & genuine                 \\ \hline
il parle de vous        & he 's talking about you & he talks about you       & he speaks of you        \\ \hline
evanouis - toi        & get out of here       & get away from it       & evacuate yourself     \\ \hline
merdique                & crappy                  & a mess                   & merchandical            \\ \hline
halte                   & stop                    & stop                     & halt                    \\ \hline
rester prudent          & be careful              & be careful               & remain cautious         \\ \hline
int{\'e}rieur               & inside                  & inside                   & within                  \\ \hline
je vous rembourserai  & i 'll pay you back    & i will pay you back    & i shall reimburse you \\ \hline
\end{tabular}
\caption{\small Examples of different wording in French-English style transfer. The French \textit{salut} translates to \textit{hey} in OpenSubtitles and to \textit{hi} in both Europarl and JRC-Acquis. \textit{evanouis-toi} can become \textit{get out of here}, \textit{get away from it} or \textit{evacuate yourself} when formality increases.
      }
      \label{tab-frenexamples}
\end{table}

 To support our assumption that the model at least somewhat consistently uses synonyms to discriminate between styles, we show in Tables \ref{tab-deenbleumet} and \ref{tab-frenbleumet} the BLEU and METEOR scores for test sets translated, respectively, from German and French into English in all style directions.

\begin{table}[]
\centering
\begin{tabular}{|c|c|c|c|}
\hline
\multirow{2}{*}{$S_{src}$} & \multicolumn{3}{c|}{$S_{tgt}$}                                  \\ \cline{2-4} 
                              & OS                   & EP                   & JRC                  \\ \hline
OS                            & \textbf{33.1/30.5} & 26.2/27.4          & 24.6/26.8          \\ \hline
EP                            & 35.4/35.9          & \textbf{38.6/37.4} & 37.7/37.0          \\ \hline
JRC                           & 52.6/42.9          & 55.4/44.2          & \textbf{58.9/45.9} \\ \hline
\end{tabular}
\caption{\small BLEU / METEOR scores of test sets translated from German into English in all style directions. METEOR typically sees a smaller decrease that BLEU, presumably due to use of style-appropriate synonyms.
      }
\label{tab-deenbleumet}
\end{table}

\begin{table}[]
\centering
\begin{tabular}{|c|c|c|c|}
\hline
\multirow{2}{*}{$S_{src}$} & \multicolumn{3}{c|}{$S_{tgt}$}                                  \\ \cline{2-4} 
                              & OS                   & EP                   & JRC                  \\ \hline
OS                            & \textbf{33.3/31.0} & 25.3/27.1          & 24.9/27.0          \\ \hline
EP                            & 39.6/38.2          & \textbf{42.4/39.6} & 41.1/39.1          \\ \hline
JRC                           & 56.6/45.3          & 59.1/46.5          & \textbf{62.6/48.1} \\ \hline
\end{tabular}
\caption{\small BLEU / METEOR scores of test sets translated from French into English in all style directions. METEOR typically sees a smaller decrease that BLEU, presumably due to use of style-appropriate synonyms.
      }
\label{tab-frenbleumet}
\end{table}

To qualitatively assess the performance of our system, we train convolutional neural network (CNN) classifiers to predict the styles of sentences presented to it, as described by ~\citet{cnn-text-classification}. We use an implementation of convolutional neural networks for text classification\footnote{\url{https://github.com/dennybritz/cnn-text-classification-tf}}. We train three separate two-class classifiers for English sentences, each of which aims to distinguish between one of the styles and the other two. The CNN classifiers are trained on a portion of the data used for training the NMT model, with filter sizes 3, 4 and 5, and 128 filters per size, with max-pooling, and dropout probability of 0.5. The classifiers for OpenSubtitles, Europarl and JRC-Acquis achieve validation accuracy of 96\%, 95\% and 96\% respectively.

Tables \ref{tab-deencnn} and \ref{tab-frencnn} demonstrate, for German-English and French-English translations respectively, the percentage of sentences classified as the target class in the human-translated test sets (that is, belonging effectively to the source class) and in the test sets translated automatically into the target class.

\begin{table}[]
\begin{tabular}{|c|c|c|c|}
\hline
\multirow{2}{*}{$S_{src}$} & \multicolumn{3}{c|}{$S_{tgt}$}             \\ \cline{2-4} 
                        & OS          & EP          & JRC         \\ \hline
OS                      & \textbf{96.4 / 97.4} & 8.1 / 24.3  & 3.4 / 5.5   \\ \hline
EP                      & 4.4 / 14.0  & \textbf{96.4 / 96.1} & 5.4 / 8.5   \\ \hline
JRC                     & 1.9 / 2.8   & 6.9 / 12.8  & \textbf{96.5 / 96.8} \\ \hline
\end{tabular}
\caption{\small Percentage of sentences recognized as the target style in human-translated German-English test sets / test sets translated from German into English and into the target style by the NMT model.
      }
\label{tab-deencnn}
\end{table}

\begin{table}[]
\centering
\begin{tabular}{|c|c|c|c|}
\hline
\multirow{2}{*}{$S_{src}$} & \multicolumn{3}{c|}{$S_{tgt}$}             \\ \cline{2-4} 
                        & OS          & EP          & JRC         \\ \hline
OS                      & \textbf{96.0 / 96.5} & 8.7 / 25.4  & 3.2 / 6.7   \\ \hline
EP                      & 3.9 / 14.1  & \textbf{95.3 / 96.2} & 4.5 / 6.4   \\ \hline
JRC                     & 1.9 / 3.7   & 7.0 / 12.4  & \textbf{96.8 / 96.7} \\ \hline
\end{tabular}
\caption{\small Percentage of sentences recognized as the target style in human-translated French-English test sets / test sets translated from French into English and into the target style by the NMT model.
      }
\label{tab-frencnn}
\end{table}

\subsection{Interpretation}

 It is clear that the system chooses shortened forms when translating into OpenSubtitles, even being over-eager and incorporating more of those than in the human translations. When translating into the more formal domains, it gets rid of most contractions (see Tables \ref{tab-deencontr}, \ref{tab-frencontr}) for both language pairs.
 
 We also expect the model to use lexical variations while translating into different styles. It is clear that both BLEU and METEOR scores should fall when the sentences are translated into a different style, because human-translated references are in the style of source sentences. However, since METEOR relies on synonymy when determining matches and BLEU does not, METEOR generally decreases about twice as less as BLEU does (see Tables \ref{tab-deenbleumet-perc} and \ref{tab-frenbleumet-perc}).
 
 Regarding the classification results, there is a consistent increase in the percentage of sentences recognized as the desired style when they are translated into that style using the Transformer NMT model. The highest rise occurs when we translate between most dissimilar styles (OS and EP) and sits in range of 200-300\%. The lowest rise accordingly occurs in case of similar styles (EP, JRC). This trend is consistent across language pairs. 
 Refer to the Tables \ref{tab-deencnn-percent}, \ref{tab-frencnn-percent} for more details. 
 
\begin{table}[]
\centering
\begin{tabular}{|c|c|c|c|}
\hline
\multirow{2}{*}{$S_{src}$} & \multicolumn{3}{c|}{$S_{tgt}$}             \\ \cline{2-4} 
                        & OS          & EP          & JRC         \\ \hline
OS                      & 1\% & 200\%  & 62\%   \\ \hline
EP                      & 218\%  & -0.3\% & 57\%   \\ \hline
JRC                     & 47\%   & 86\%  & 0.3\% \\ \hline
\end{tabular}
\caption{\small Relative change in style. Percentages show how much the amount of the target style in corpus increased after cross-lingual zero-shot style transfer. Language direction: German into English}
\label{tab-deencnn-percent}
\end{table}

\begin{table}[]
\centering
\begin{tabular}{|c|c|c|c|}
\hline
\multirow{2}{*}{$S_{src}$} & \multicolumn{3}{c|}{$S_{tgt}$}             \\ \cline{2-4} 
                        & OS          & EP          & JRC         \\ \hline
OS                      & 0.5\% & 192\%  & 109\%   \\ \hline
EP                      & 262\%  & 0.9\% & 42\%   \\ \hline
JRC                     & 95\%   & 77\%  & 0.1\% \\ \hline
\end{tabular}
\caption{\small Relative change in style. Percentages show how much the amount of the target style in corpus increased after cross-lingual zero-shot style transfer. Language direction: French into English.
      }
\label{tab-frencnn-percent}
\end{table}

\begin{table}[]
\centering
\begin{tabular}{|c|c|c|c|}
\hline
\multirow{2}{*}{$S_{src}$} & \multicolumn{3}{c|}{$S_{tgt}$}                                  \\ \cline{2-4} 
                              & OS                   & EP                   & JRC                  \\ \hline
OS                            & \textbf{-} & 20.9/10.2\%          & 25.7/12.1\%          \\ \hline
EP                            & 8.3/4.0\%          & \textbf{-} & 2.3/1.1\%          \\ \hline
JRC                           & 10.7/6.5\%          & 5.9/3.7\%          & \textbf{-} \\ \hline
\end{tabular}
\caption{\small Relative decrease in BLEU / METEOR scores of test sets translated from German into English in all style directions. METEOR typically drops about twice as less comparing to the BLEU, presumably due to use of style-appropriate synonyms.
      }
\label{tab-deenbleumet-perc}
\end{table}

\begin{table}[]
\centering
\begin{tabular}{|c|c|c|c|}
\hline
\multirow{2}{*}{$S_{src}$} & \multicolumn{3}{c|}{$S_{tgt}$}                                  \\ \cline{2-4} 
                              & OS                   & EP                   & JRC                  \\ \hline
OS                            & \textbf{-} & 24.0/12.6\%          & 25.3/12.9\%          \\ \hline
EP                            & 6.6/3.5\%          & \textbf{-} & 3.1/1.3\%        \\ \hline
JRC                           & 9.6/5.8\%          & 5.6/3.3\%          & \textbf{-} \\ \hline
\end{tabular}
\caption{\small Relative decrease in BLEU / METEOR scores of test sets translated from French into English in all style directions. METEOR typically drops about twice as less comparing to the BLEU, presumably due to use of style-appropriate synonyms.
      }
\label{tab-frenbleumet-perc}
\end{table}

\subsection{Monolingual Zero-shot Style Transfer}
While the main focus of our paper has been so far on cross-lingual style transfer, as a side effect, translating from English to English while passing different target styles to the model potentially leads to a monolingual (zero-shot) style transfer. Tables \ref{tab-enenbleumet} and \ref{tab-enencontr} demonstrate that this is indeed the case and trends observed in cross-lingual transfer reappear.

\begin{table}[]
\centering
\begin{tabular}{|c|c|c|c|}
\hline
\multirow{2}{*}{$S_{src}$} & \multicolumn{3}{c|}{$S_{tgt}$}                                  \\ \cline{2-4} 
                              & OS                   & EP                   & JRC                  \\ \hline
OS                            & \textbf{85.4/60.9} & 70.0/50.3          & 75.0/53.7          \\ \hline
EP                            & 85.5/61.3          & \textbf{83.6/60.4} & 82.1/59.4          \\ \hline
JRC                           & 90.9/64.7          & 87.9/62.1          & \textbf{87.9/61.9} \\ \hline
\end{tabular}
\caption{\small BLEU / METEOR scores of test sets translated from English into English in all style directions.
      }
\label{tab-enenbleumet}
\end{table}

\begin{table}[]
\centering
\begin{tabular}{|c|c|c|c|}
\hline
\multirow{2}{*}{$S_{src}$} & \multicolumn{3}{c|}{$S_{tgt}$}\\
\cline{2-4} 
                      & OS                 & EP             & JRC            \\ \hline
OS                    & \textbf{390 (363)} & 23             & 70             \\ \hline
EP                    & 167                & \textbf{0 (0)} & 0              \\ \hline
JRC                   & 4                 & 0              & \textbf{0 (0)} \\ \hline
\end{tabular}
\caption{\small Number of contractions in 1000-sentence test sets when translated from English into English in all 9 style directions. The numbers in parentheses indicate the number of contracted forms in the human-translated test sets.
      }
      \label{tab-enencontr}
\end{table}

\section{Discussion}
\label{sec:discussion}

The proportion of sentences recognized as members of the desired target classes by the CNN domain classifiers remains quite low. However, we believe that such evaluation may be less suitable for our task than it is for the more typical tasks in style transfer, such as sentiment or gender transfer.

Our CNN network is, in fact, a domain classifier,
trained, unavoidably in our case, to classify relying not only on style, but on semantic content as well. Thus, the topic of a sentence might significantly influence the classifier decision. For example, it is highly unlikely to have words typical for such genres as \textit{fantasy} or \textit{rock'n'roll} in Europarl (European Parliament speech transcripts) corpus, while it is natural to expect such topics from OpenSubtitles texts. The style transfer itself can influence some lexical choices and the general feeling of the sentence, but not dramatically enough to fool our domain classification model on vast majority of cases. The topics differ significantly between the corpora we use, and semantic content should greatly influence classifier output.

From this perspective, our style transfer model is strong enough to make the domain classification model increase its predictions of the target domain by up to three times (Tables \ref{tab-deencnn-percent}, \ref{tab-frencnn-percent}). We interpret it as a solid result.

It should be also noted that while we only provide results for English as the target language for now, the system is, by design, multilingual, and there is nothing English-specific to it. All assumptions should hold true for other target languages as well. We chose English as a target language to simplify evaluation and understanding.

\section{Related Work}
\label{sec:related-work}

The task of cross-lingual zero-shot style transfer is novel and is not discussed in the literature at the best of our knowledge. However, many recent models for style transfer that do not rely on direct parallel signal contain innate structure that is aimed to separate content and style representations \cite{gvae-1, gvae-2, gvae-3}. This is often done using VAE's \cite{vae} and GAN's \cite{GAN}, while sometimes explicit attribute substitutions are involved \cite{manual}.

We in turn employ ideas from multilingual NMT systems \cite{multiling-multiway, multiling-goolge}, which are proven to have a high performance while also enabling translation between language pairs that we not have direct parallel corpora. We use a similar multilingual architecture while also extending its benefits to styles. 

Several authors also employed machine translation ideas to support style transfer task. Approaches include back-translating languages \cite{mt-backtrans} and using seq2seq architectures \cite{mt-seq2seq} as a part of the pipeline. 

Most similar work to ours conceptually is done by \citet{rnns-zeroshot}. They aim to do monolingual zero-shot style transfer, and use 32 different English versions of Bible as their source of paraphrase training data. We use multiple languages, multiple styles, and do not rely on any explicit lexicons. 

Finally, \citet{formality} inspect generating NMT output with different degrees of formality. Methods include vector space models that require large mixed-topic corpora to be trained. While being similar in a way that it also modifies NMT output, the main focus of our work is cross-lingual zero-shot style transfer. Moreover, our work is not limited to generating hypotheses with different degrees formality, but arbitrary styles can be used.

\section{Conclusion}
\label{sec:conclusion}

We propose a method for cross-lingual zero-shot style transfer. While our model does not achieve high scores based on classifier output, we believe that such evaluation may be less suitable for our tasks than it is for the more typical tasks in style transfer, such as sentiment or gender transfer, because the topics differ significantly between the corpora we use, and semantic content should greatly influence classifier output.

However, the model does show promising qualitative results, demonstrating the ability to capture some important aspects of stylistic difference between domains.

\section{Future work}
\label{sec:future}

In the future, we plan to assess other aspects of style differences that the model may capture, and to use help of human evaluators in assessing our results. We intend to perform a meaningful qualitative and quantitative comparison to a previously existing strong style transfer system. We also intend to train and evaluate two-language NMT systems with the same approach to learning style to see how multilinguality interferes with style transfer.

In addition, we plan to improve quantitative evaluation of the model by making the CNN classifiers more appropriate for the task, for instance, by training them as multi-class and yielding class probabilities rather than labels, and by making them more content-agnostic and style-sensitive.

\section*{Acknowledgments}

The authors would like to thank the University for providing GPU computing resources. \\

\bibliography{emnlp2018}

\begin{thebibliography}{23}
\expandafter\ifx\csname natexlab\endcsname\relax\def\natexlab#1{#1}\fi

\bibitem[{Banerjee and Lavie(2005)}]{meteor}
Satanjeev Banerjee and Alon Lavie. 2005.
\newblock {METEOR}: An automatic metric for mt evaluation with improved
  correlation with human judgments.
\newblock In \emph{Proceedings of the acl workshop on intrinsic and extrinsic
  evaluation measures for machine translation and/or summarization}, pages
  65--72.

\bibitem[{Carlson et~al.(2017)Carlson, Riddell, and Rockmore}]{rnns-zeroshot}
Keith Carlson, Allen Riddell, and Daniel~N. Rockmore. 2017.
\newblock Zero-shot style transfer in text using recurrent neural networks.
\newblock \emph{CoRR}, abs/1711.04731.

\bibitem[{Firat et~al.(2016)Firat, Cho, and Bengio}]{multiling-multiway}
Orhan Firat, KyungHyun Cho, and Yoshua Bengio. 2016.
\newblock Multi-way, multilingual neural machine translation with a shared
  attention mechanism.
\newblock \emph{CoRR}, abs/1601.01073.

\bibitem[{{Goodfellow} et~al.(2014){Goodfellow}, {Pouget-Abadie}, {Mirza},
  {Xu}, {Warde-Farley}, {Ozair}, {Courville}, and {Bengio}}]{GAN}
I.~J. {Goodfellow}, J.~{Pouget-Abadie}, M.~{Mirza}, B.~{Xu}, D.~{Warde-Farley},
  S.~{Ozair}, A.~{Courville}, and Y.~{Bengio}. 2014.
\newblock {Generative Adversarial Networks}.
\newblock \emph{ArXiv e-prints}.

\bibitem[{Han et~al.(2018)Han, Wu, and Niu}]{mt-seq2seq}
Mengqiao Han, Ou~Wu, and Zhendong Niu. 2018.
\newblock Unsupervised automatic text style transfer using lstm.
\newblock pages 281--292.

\bibitem[{Hieber et~al.(2017)Hieber, Domhan, Denkowski, Vilar, Sokolov,
  Clifton, and Post}]{sockeye}
Felix Hieber, Tobias Domhan, Michael Denkowski, David Vilar, Artem Sokolov, Ann
  Clifton, and Matt Post. 2017.
\newblock Sockeye: {A} toolkit for neural machine translation.
\newblock \emph{CoRR}, abs/1712.05690.

\bibitem[{Hu et~al.(2017)Hu, Yang, Liang, Salakhutdinov, and Xing}]{gvae-1}
Zhiting Hu, Zichao Yang, Xiaodan Liang, Ruslan Salakhutdinov, and Eric~P. Xing.
  2017.
\newblock Controllable text generation.
\newblock \emph{CoRR}, abs/1703.00955.

\bibitem[{Johnson et~al.(2016)Johnson, Schuster, Le, Krikun, Wu, Chen, Thorat,
  Vi{\'{e}}gas, Wattenberg, Corrado, Hughes, and Dean}]{multiling-goolge}
Melvin Johnson, Mike Schuster, Quoc~V. Le, Maxim Krikun, Yonghui Wu, Zhifeng
  Chen, Nikhil Thorat, Fernanda~B. Vi{\'{e}}gas, Martin Wattenberg, Greg
  Corrado, Macduff Hughes, and Jeffrey Dean. 2016.
\newblock Google's multilingual neural machine translation system: Enabling
  zero-shot translation.
\newblock \emph{CoRR}, abs/1611.04558.

\bibitem[{Kim(2014)}]{cnn-text-classification}
Yoon Kim. 2014.
\newblock Convolutional neural networks for sentence classification.
\newblock \emph{CoRR}, abs/1408.5882.

\bibitem[{{Kingma} and {Welling}(2013)}]{vae}
D.~P {Kingma} and M.~{Welling}. 2013.
\newblock {Auto-Encoding Variational Bayes}.
\newblock \emph{ArXiv e-prints}.

\bibitem[{Koehn(2005)}]{europarl}
Philipp Koehn. 2005.
\newblock Europarl: A parallel corpus for statistical machine translation.
\newblock In \emph{MT summit}, volume~5, pages 79--86.

\bibitem[{Koehn et~al.(2007)Koehn, Hoang, Birch, Callison-burch, Zens,
  Federico, Bertoldi, Dyer, Cowan, Shen, Moran, Bojar, Constantin, and
  Herbst}]{moses}
Philipp Koehn, Hieu Hoang, Alexandra Birch, Chris Callison-burch, Richard Zens,
  Marcello Federico, Nicola Bertoldi, Chris Dyer, Brooke Cowan, Wade Shen,
  Christine Moran, Ondrej Bojar, Alexandra Constantin, and Evan Herbst. 2007.
\newblock Moses: Open source toolkit for statistical machine translation.

\bibitem[{Li et~al.(2018)Li, Jia, He, and Liang}]{manual}
Juncen Li, Robin Jia, He~He, and Percy Liang. 2018.
\newblock Delete, retrieve, generate: {A} simple approach to sentiment and
  style transfer.
\newblock \emph{CoRR}, abs/1804.06437.

\bibitem[{Lison and Tiedemann(2016)}]{opensubs}
Pierre Lison and Jörg Tiedemann. 2016.
\newblock Opensubtitles2016: Extracting large parallel corpora from movie and
  tv subtitles.
\newblock In \emph{Proceedings of the Tenth International Conference on
  Language Resources and Evaluation (LREC 2016)}, Paris, France. European
  Language Resources Association (ELRA).

\bibitem[{Mueller et~al.(2017)Mueller, Gifford, and Jaakkola}]{gvae-2}
Jonas Mueller, David Gifford, and Tommi Jaakkola. 2017.
\newblock Sequence to better sequence: Continuous revision of combinatorial
  structures.
\newblock In \emph{Proceedings of the 34th International Conference on Machine
  Learning}, volume~70 of \emph{Proceedings of Machine Learning Research},
  pages 2536--2544, International Convention Centre, Sydney, Australia. PMLR.

\bibitem[{Niu et~al.(2017)Niu, Martindale, and Carpuat}]{formality}
Xing Niu, Marianna~J. Martindale, and Marine Carpuat. 2017.
\newblock A study of style in machine translation: Controlling the formality of
  machine translation output.

\bibitem[{Papineni et~al.(2002)Papineni, Roukos, Ward, and Zhu}]{bleu}
Kishore Papineni, Salim Roukos, Todd Ward, and Wei-Jing Zhu. 2002.
\newblock Bleu: a method for automatic evaluation of machine translation.
\newblock In \emph{Proceedings of 40th Annual Meeting of the Association for
  Computational Linguistics}, pages 311--318, Philadelphia, Pennsylvania, USA.
  Association for Computational Linguistics.

\bibitem[{Prabhumoye et~al.(2018)Prabhumoye, Tsvetkov, Salakhutdinov, and
  Black}]{mt-backtrans}
Shrimai Prabhumoye, Yulia Tsvetkov, Ruslan Salakhutdinov, and Alan~W Black.
  2018.
\newblock Style transfer through back-translation.

\bibitem[{Sennrich and Haddow(2016)}]{factors}
Rico Sennrich and Barry Haddow. 2016.
\newblock Linguistic input features improve neural machine translation.
\newblock \emph{CoRR}, abs/1606.02892.

\bibitem[{Shen et~al.(2017)Shen, Lei, Barzilay, and Jaakkola}]{gvae-3}
Tianxiao Shen, Tao Lei, Regina Barzilay, and Tommi~S. Jaakkola. 2017.
\newblock Style transfer from non-parallel text by cross-alignment.
\newblock \emph{CoRR}, abs/1705.09655.

\bibitem[{Steinberger et~al.(2006)Steinberger, Pouliquen, Widiger, Ignat,
  Erjavec, Tufi{\c{s}}, and Varga}]{jrc}
Ralf Steinberger, Bruno Pouliquen, Anna Widiger, Camelia Ignat, Toma\v{z}
  Erjavec, Dan Tufi{\c{s}}, and D\'{a}niel Varga. 2006.
\newblock The jrc-acquis: A multilingual aligned parallel corpus with 20+
  languages.
\newblock In \emph{Proceedings of the Fifth International Conference on
  Language Resources and Evaluation (LREC'06)}.

\bibitem[{Vaswani et~al.(2017)Vaswani, Shazeer, Parmar, Uszkoreit, Jones,
  Gomez, Kaiser, and Polosukhin}]{transformer}
Ashish Vaswani, Noam Shazeer, Niki Parmar, Jakob Uszkoreit, Llion Jones,
  Aidan~N. Gomez, Lukasz Kaiser, and Illia Polosukhin. 2017.
\newblock Attention is all you need.
\newblock \emph{CoRR}, abs/1706.03762.

\bibitem[{Wu et~al.(2016)Wu, Schuster, Chen, Le, Norouzi, Macherey, Krikun,
  Cao, Gao, Macherey, Klingner, Shah, Johnson, Liu, Kaiser, Gouws, Kato, Kudo,
  Kazawa, Stevens, Kurian, Patil, Wang, Young, Smith, Riesa, Rudnick, Vinyals,
  Corrado, Hughes, and Dean}]{googleSentencePieces}
Yonghui Wu, Mike Schuster, Zhifeng Chen, Quoc~V. Le, Mohammad Norouzi, Wolfgang
  Macherey, Maxim Krikun, Yuan Cao, Qin Gao, Klaus Macherey, Jeff Klingner,
  Apurva Shah, Melvin Johnson, Xiaobing Liu, Lukasz Kaiser, Stephan Gouws,
  Yoshikiyo Kato, Taku Kudo, Hideto Kazawa, Keith Stevens, George Kurian,
  Nishant Patil, Wei Wang, Cliff Young, Jason Smith, Jason Riesa, Alex Rudnick,
  Oriol Vinyals, Greg Corrado, Macduff Hughes, and Jeffrey Dean. 2016.
\newblock Google's neural machine translation system: Bridging the gap between
  human and machine translation.
\newblock \emph{CoRR}, abs/1609.08144.

\end{thebibliography}
\bibliographystyle{acl_natbib_nourl}

\appendix

\section{Supplemental Material}
\label{sec:supplemental}

\begin{table*}[]
\centering
\begin{tabular}{|p{3.5cm}|p{3.5cm}|p{3.5cm}|p{3.5cm}|}
\hline
\textbf{Source (de)}                             & \textbf{OS}                                & \textbf{EP}                                          & \textbf{JRC}                                   \\ \hline
m{\"o}chte ich                              & i want                            & i would like                                & i would like                          \\ \hline
erst                                    & first                             & first of all                                & first of all                          \\ \hline
damit du dich hinlegen kannst           & so you can lie down               & so that you can lie down                    & and lie down                          \\ \hline
er jetzt verr{\"u}ckt wird                  & he's going crazy now              & he gets mad now                             & he becomes crazy now                  \\ \hline
schnell                                 & fast                              & quickly                                     & quickly                               \\ \hline
ich reise nach london weiter            & i'm going to london               & i will continue to travel to london         & i shall continue to travel to london  \\ \hline
aber                                    & but                               & however                                     & however                               \\ \hline
ich wechsle                             & i'm turning                       & i am turning                                & i shall refer                         \\ \hline
mach das nicht                          & don't do that                     & don't do that                               & do not do so                          \\ \hline
tats{\"a}chlich                             & actually                          & indeed                                      & indeed                                \\ \hline
wir gehen                               & we're leaving                     & we are leaving                              & we shall leave                        \\ \hline
wer                                     & whoever                           & those who                                   & those who                             \\ \hline
ich halte es auf                        & i'll stop it                      & i will stop it                              & i shall stop it                       \\ \hline
nichts derartiges                       & anything like that                & anything like this                          & anything of this kind                 \\ \hline
ich h{\"a}tte gedacht                       & i thought                         & i would have thought                        & i would have thought                  \\ \hline
sonst                                   & or                                & otherwise                                   & otherwise                             \\ \hline
um den schmerz zu lindern               & to ease the pain                  & in order to alleviate the pain              & in order to alleviate the pain        \\ \hline
noch viel mehr                          & more that than                    & much more                                   & even more so                          \\ \hline
ja                                      & yeah                              & yes                                         & yes                                   \\ \hline
gr{\"o}{\ss}er                                  & bigger                            & greater                                     & greater                               \\ \hline
eine chance                             & a chance                          & an opportunity                              & an opportunity                        \\ \hline
auch der M{\"o}rder                         & the killer, too                   & also the murderer                           & also the murderer                     \\ \hline
heutzutage ist es schwerer              & nowadays, it's harder             & today, it is more difficult                 & it is now more difficult              \\ \hline
wie geht's dann weiter?                 & so, what's next?                  & what happens then?                          & how are we to proceed then?           \\ \hline
ein r{\"a}tsel, das es zu l{\"o}sen gilt        & a mystery to solve                & a mystery that needs to be resolved         & a mystery to be resolved              \\ \hline
bleibt                                  & stay                              & remain                                      & remain                                \\ \hline
wie schon gesagt                        & like i said                       & as i have already said                      & as already stated                     \\ \hline
ich mache eine anfrage                  & i'll ask you a question           & i have a question                           & i shall make a request                \\ \hline
ich wollte sie nicht kr{\"a}nken            & i didn't mean to hurt you         & i did not want to offend you                & i did not wish to offend you          \\ \hline
damit ich das richtig verstehe          & let me get this straight          & let me get this straight                    & in order to understand this correctly \\ \hline
sie haben die fotos                     & they've got the photos            & they have the photographs                   & they shall have the photographs       \\ \hline
shauen sie genau hin                    & look carefully                    & look carefully                              & take a close look                     \\ \hline
du f{\"u}r kommunikation zust{\"a}ndig bist     & you're in charge of communication & you are responsible for communication       & you are responsible for communication \\ \hline
ich wagte mich schon zu weit vor        & i've gone too far                 & i have dared to go too far                  & i have dared to go too far            \\ \hline
ohne auch nur kurz dar{\"u}ber nachzudenken & without even thinking about it    & without even thinking a little bit about it & without even considering it briefly   \\ \hline
au{\ss}erdem                                & besides                           & moreover                                    & moreover                              \\ \hline
ach ja?                                 & oh, yeah?                         & is that so?                                 & is that so?                           \\ \hline
\end{tabular}
\caption{\small Additional examples of difference in lexical and grammatical choices when translating from German to English into different styles.
      }
      \label{tab-deen-examples-extended}
\end{table*}

\begin{table*}[]
\begin{tabular}{|p{3.5cm}|p{3.5cm}|p{3.5cm}|p{3.5cm}|}
\hline
\textbf{Source (fr)}                                                                                                                                            & \textbf{OS}                                                                                                                           & \textbf{EP}                                                                                                                                                  & \textbf{JRC}                                                                                                                                               \\ \hline
il para{\^i}t que vous {\^e}tes sp{\'e}ciale .                                                                                                                & i hear you 're special .                                                                                                     & i understand that you are special .                                                                                                                 & it appears that you are special .                                                                                                                 \\ \hline
c'est {\c c}a                                                                                                                                        & that 's right                                                                                                              & that is it                                                                                                                                      & that is the case                                                                                                                                \\ \hline
tr{\`e}s vilaine                                                                                                                                    & very bad                                                                                                                   & very bad                                                                                                                                          & very vile                                                                                                                                       \\ \hline
vrai                                                                                                                                              & real                                                                                                                         & genuine                                                                                                                                             & genuine                                                                                                                                           \\ \hline
on a un autre concert demain                                                                                                                      & we 've got another concert tomorrow .                                                                                        & we have another concert tomorrow .                                                                                                                  & a further concert will be held tomorrow                                                                                                           \\ \hline
on se tire une balle                                                                                                                              & we shoot each other                                                                                                          & you get a bullet                                                                                                                                    & a bullet is fired                                                                                                                                 \\ \hline
donc                                                                                                                           & so                                                                                                     & so                                                                                                                        & therefore                                                                                                                \\ \hline
bien , on a fini .                                                                                                                                & all right , we 're done .                                                                                                    & well , we have finished .                                                                                                                           & well , we have finished .                                                                                                                         \\ \hline
qu'est - ce que vous mijotez?                                                                                                                     & what are you up to ?                                                                                                         & what are you doing ?                                                                                                                                & what are you doing ?                                                                                                                              \\ \hline
manquent de respect                                                                                                                               & have no respect                                                                                                              & do not respect                                                                                                                                      & are disrespectful to                                                                                                                              \\ \hline
vous m'aviez indiqu{\'e}                                                     & you told me                                                   & you indicated to me                                                                 & you indicated to me                                                               \\ \hline
enl{\`e}ve l'argent                                                                                                                     & take the money away                                                                                       & take the money away                                                                                                              & remove the money                                                                                                              \\ \hline
rester prudent                                                                                                                                    & be careful                                                                                                                   & be careful                                                                                                                                          & remain cautious                                                                                                                                   \\ \hline
une r{\'e}union est organis{\'e}e {\`a} la mairie demain soir , - si vous souhaitez ...                                                                       & there 's a meeting at the city hall tomorrow night , if you 'd like . . .                                                    & a meeting is being held in the city hall tomorrow evening - if you wish . . .                                                                       & a meeting shall be held at the council meeting tomorrow evening , - if you wish . . .                                                             \\ \hline
on l'a d{\'e}tect{\'e}e                                                                                                                                   & we detected it                                                                                                               & it was detected                                                                                                                                     & it was detected                                                                                                                                   \\ \hline
deux heures                                                                                                                           & two hours                                                                                                                    & two hours                                                                                                                                           & a two-hour period                                                                                                                                 \\ \hline
pr{\'e}cieuse                                                                                                                                         & precious                                                                                                                     & valuable                                                                                                                                            & valuable                                                                                                                                          \\ \hline
vous ne voyez pas qu'il a r{\'e}pondu {\`a} une provocation ?                                                                                             & can 't you see he responded to a provocation ?                                                                               & do you not see that he responded to a provocation ?                                                                                                 & do you not see that he responded to a provocation ?                                                                                               \\ \hline
c'est incroyable !                                                                                                                                & it 's amazing !                                                                                                              & that is unbelievable !                                                                                                                              & this is unbelievable !                                                                                                                            \\ \hline
halte                                                                                                                                             & stop                                                                                                                         & stop                                                                                                                                                & halt                                                                                                                                              \\ \hline
ca risque d'{\^e}tre une piste difficile                                                                                                            & it 's gonna be a hard run                                                                                                  & there is a risk that this will be a difficult path                                                                                                & this is likely to be a difficult path                                                                                                           \\ \hline
{\c c}a m'a fait plaisir de le voir heureux                                                                                                          & it made me happy to see him happy                                                                                          & i was pleased to see him happy                                                                                                                   & i was pleased to see him happy                                                                                                                  \\ \hline
le petit a disparu & the kid 's gone & the child has disappeared & the child has disappeared \\ \hline
film                                                                                                                                              & movie                                                                                                                        & film                                                                                                                                                & film                                                                                                                                              \\ \hline
journaux                                                                                                                                          & papers                                                                                                                       & newspapers                                                                                                                                          & newspapers                                                                                                                                        \\ \hline
merdique                                                                                                                                          & crappy                                                                                                                       & a mess                                                                                                                                              & merchandical                                                                                                                                      \\ \hline
je vous rembourserai .                                                                                                                            & i 'll pay you back .                                                                                                         & i will pay you back .                                                                                                                               & i shall reimburse you .                                                                                                                           \\ \hline
procureur                                                                                                                                         & d.a.                                                                                                                         & prosecutor                                                                                                                                          & public prosecutor                                                                                                                                 \\ \hline
personne                                                                                                                                          & no one                                                                                                                     & nobody                                                                                                                                           & no person                                                                                                                                       \\ \hline
honte                                                                                                                                             & shame                                                                                                                        & disgrace                                                                                                                                            & disgrace                                                                                                                                          \\ \hline
evanouis - toi .                                                                                                                                  & get out of here .                                                                                                            & get away from it .                                                                                                                                  & evacuate yourself .                                                                                                                               \\ \hline
il parle de vous                                                                                                                                  & he 's talking about you                                                                                                      & he talks about you                                                                                                                                  & he speaks of you                                                                                                                                  \\ \hline
on se lance?                                                                                                                                      & let 's go .                                                                                                                  & are we getting started ?                                                                                                                            & are we going ?                                                                                                                                    \\ \hline
salut                                                                                                                                             & hey                                                                                                                          & hi                                                                                                                                                  & hi                                                                                                                                                \\ \hline
on me prendrait pour un idiot                                                                                                                  & they 'd think i 'm an idiot                                                                                                & i would be thought to be an idiot                                                                                                                 & i would be regarded as an idiot                                                                                                                 \\ \hline
\end{tabular}
\caption{\small Additional examples of difference in lexical and grammatical choices when translating from French to English into different styles.
      }
\end{table*}

\end{document}